\documentclass{article}

\usepackage[final]{corl_2020} 
\usepackage{amsmath,amsfonts,amsthm,amssymb}
\usepackage{bbm}
\usepackage{graphicx}
\usepackage{wrapfig}
\usepackage{floatrow}
\usepackage{subcaption}
\usepackage{algorithm}
\usepackage{algpseudocode}

\usepackage{siunitx}
\newtheoremstyle{nospace}
{2pt}   				
{2pt}   				
{\itshape}  			
{} 		  		    	
{\bfseries} 			
{.}         			
{5pt plus 1pt minus 1pt}
{}          			

\theoremstyle{nospace} \newtheorem{definition}{Definition}

\usepackage{titlesec}
\titlespacing*{\section}{0pt}{0.5ex plus 0.1ex minus .1ex}{0.5ex plus .1ex}
\titlespacing*{\subsection}{0pt}{0.1ex plus 1ex minus .2ex}{0.1ex plus .2ex}

\title{Learning to Actively Reduce Memory Requirements for Robot Control Tasks}

\author{
  Meghan Booker and Anirudha Majumdar\\
  Department of Mechanical and Aerospace Engineering\\
  Princeton University\\
  \texttt{\{mebooker, ani.majumdar\}@princeton.edu} \\
}

\begin{document}
\maketitle


\begin{abstract}
Robots equipped with rich sensing modalities (e.g., RGB-D cameras) performing long-horizon tasks motivate the need for policies that are highly \emph{memory-efficient}. 
State-of-the-art approaches for controlling robots often use memory representations that are excessively rich for the task or rely on hand-crafted tricks for memory efficiency. 
Instead, this work provides a general approach for \emph{jointly synthesizing} memory representations and policies; the resulting policies \emph{actively seek} to reduce memory requirements.
Specifically, we present a reinforcement learning framework that leverages an implementation of the \emph{group LASSO} regularization to synthesize policies that employ low-dimensional and task-centric memory representations.  We demonstrate the efficacy of our approach with simulated examples including navigation in discrete and continuous spaces as well as vision-based indoor navigation set in a photo-realistic simulator. The results on these examples indicate that our method is capable of finding policies that rely only on low-dimensional memory representations, improving generalization, and actively reducing memory requirements. 
\end{abstract}

\keywords{Memory-Efficiency, Navigation, Reinforcement Learning} 

\section{Introduction}
\label{sec:introduction}
Consider a robot given a coverage task on a building floor. For example, it could be tasked with performing a safety inspection or collecting data. With the increasing availability and use of high-resolution sensors such as cameras and LiDAR, such tasks require the robot to process high-dimensional observations for real-time decisions. Current navigation approaches typically involve constructing and utilizing a high-fidelity map of the robot's environment~\citep{Cadena16, Sun18, Doherty19, Vasilopoulos20}. However, is a map necessary for the task?  Does the map-based representation satisfy the robot's onboard memory constraints? Are there representations that are more memory efficient? These fundamental and practical questions motivate the need to have principled methods for finding memory representations that are not only sufficient for the task at hand but also reduce the robot's memory requirements.

In order to illustrate the potential benefits of memory-efficient policies, consider a robot tasked with covering an $n \times n$ maze (a simplified version of the building floor coverage task). Blum and Kozen \cite{Blum78} show that there is a control policy --- a clever, handcrafted, wall-following and zig-zagging routine --- that only utilizes $O(\log n)$ bits of memory. This policy thus requires \emph{significantly less} memory than one that relies on building and using a map of the environment (a map-building strategy requires at least $O(n^2)$ memory). Beyond memory efficiency, such a policy also affords additional important advantages including (i) computational efficiency, and (ii) improved generalization/robustness. For example, Blum and Kozen's policy does not need to perform real-time computations with the entire map as an input. Additionally, a policy that requires $O(\log n)$ memory is inherently \emph{task-centric}; irrelevant geometric details of the environment (e.g., the exact positions or colors of obstacles in the environment) do not affect the robot's behavior. The policy can thus be highly robust to uncertainty or noise in these task-irrelevant features.

\begin{figure}[t]
\begin{subfigure}{0.45\textwidth}
\includegraphics[width=0.88\columnwidth,trim=0 0 0 150, clip]{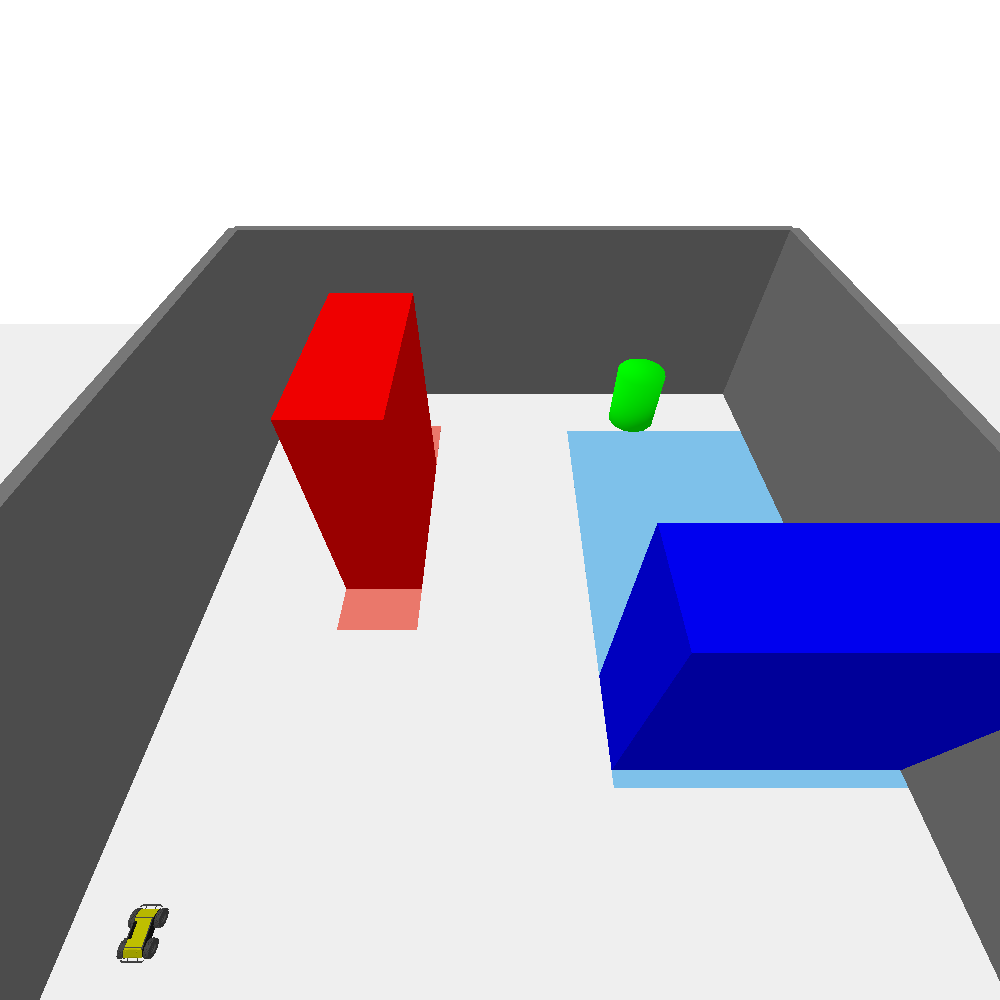}
\caption{}
\label{fig:obstacle}
\end{subfigure}
\begin{subfigure}{0.45\textwidth}
\includegraphics[width=0.88\columnwidth,trim=0 0 0 95, clip]{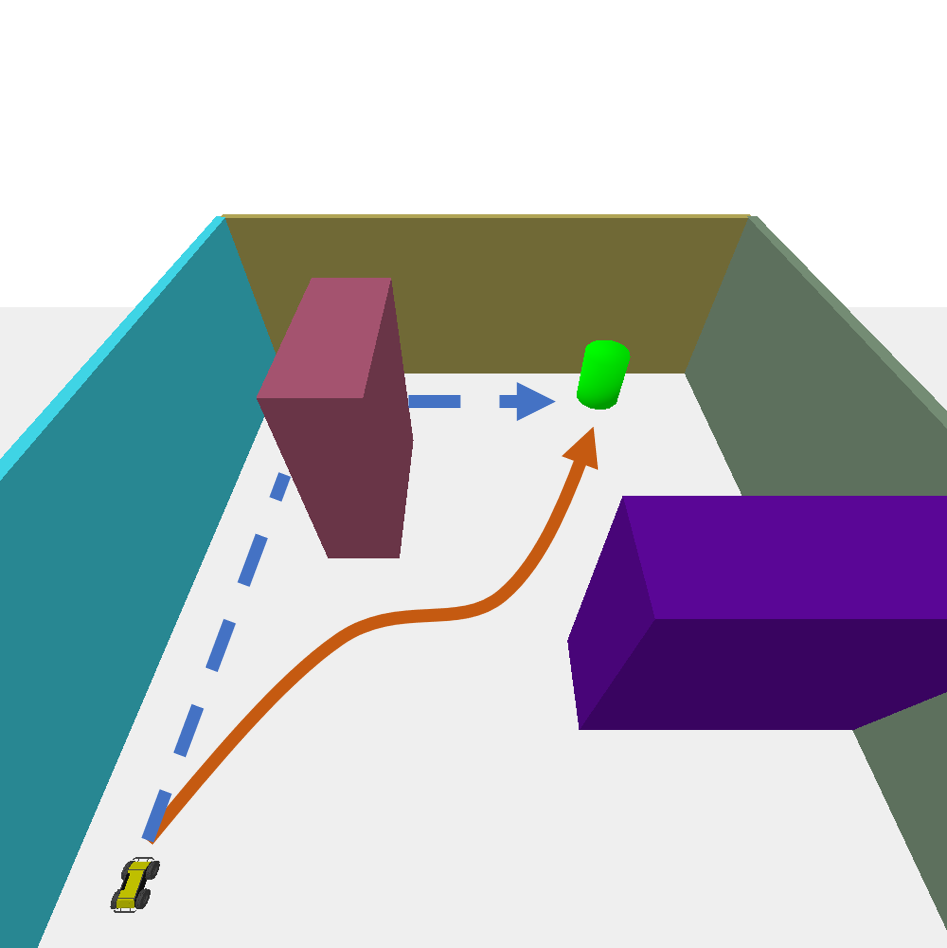}
\caption{}
\label{fig:obstacle_test}
\end{subfigure}
\vspace{-5pt}
\caption{\footnotesize{A depiction of the maze navigation problem with continuous state and action spaces described in Section \ref{mazenav}. The Husky robot needs to navigate from the southwest corner to the goal (green) in the northeast corner. \textbf{(a)} A sample maze used in training. The red and blue obstacles are placed within the respective shaded regions.  \textbf{(b)} New maze introduced during testing. The paths on the floor illustrate the policies found by a standard policy gradient (PG) method (solid orange) and the approach presented here (dotted blue). The latter (wall-following) policy is significantly more memory efficient and is also robust to changes to the distribution of obstacle colors.
}\label{fig:cont_maze}
\vspace{-15pt}
}
\end{figure}

An important feature of memory-efficient policies is that they can be \emph{qualitatively different} from ones that utilize map-based representations. As a simple example, consider the navigation problem demonstrated in Figure \ref{fig:cont_maze}. A policy that chooses to follow the wall can be significantly more memory-efficient than one that navigates through the environment diagonally (since the wall-following strategy does not need to maintain information pertaining to obstacle locations). This motivates the need to \emph{jointly synthesize} the memory representation and the control policy; such a joint synthesis can lead to policies that \emph{actively} reduce memory requirements.

\textbf{Statement of Contributions.} The goal of this paper is to synthesize low-dimensional, task-centric memory representations and control policies. Our primary contribution is a reinforcement learning framework for finding policies that achieve \emph{active memory reduction (AMR)}. In particular, we leverage a group LASSO regularization scheme~\citep{Yuan06, Scardapane17} to enforce low-dimensional memory representations while simultaneously finding policies via a policy gradient (PG)-style algorithm that we refer to as AMR-PG. To our knowledge, this is the first work to find AMR policies in continuous state and action spaces. Lastly, we demonstrate the efficacy of our approach on three simulated examples that demonstrate our method's ability to find AMR policies that reduce the dimension of the required memory representation and improve generalization as compared to standard PG methods.
\section{Related Work}
\label{sec:related}
\textbf{Memory-Efficient Representations.} There are several approaches that consider memory-efficient representations for robot navigation tasks including gap navigation trees~\citep{Murphy08, Tovar05}, compact maps~\citep{Srivastava16}, and graph-like topometric maps~\citep{Ort20}.  While each of these share this work's goal of memory efficiency, these memory representations are hand-crafted for certain applications or domains. In contrast, we aim to provide a general approach for finding memory-efficient representations and policies. 
Recent work by O'Kane and Shell~\citep{OKane17} takes a step in this direction by automatically designing minimal memory representations and policies via combinatorial filters. However, their formulation defines memory with respect to the number of nodes in a policy graph and is restricted to discretized state and action spaces. Instead, our work defines memory complexity as the dimension of a continuous representation and is applicable to continuous state and action spaces. 

\textbf{Map-Free Representations in RL.} End-to-end reinforcement learning (RL) of policies provides one avenue towards generating task-centric representations that avoid explicit geometric representations such as maps (see, e.g., \cite{Levine16, Levine18, Zhu17}). Recurrent neural network (RNN) architectures allow one to incorporate memory into policies learned via RL~\citep{Heess15}. For example, in the context of navigation, Chen et al.~\citep{Chen17} use a long short-term memory (LSTM)~\citep{Hochreiter97} architecture to navigate mazes with cul-de-sacs. While these approaches are able to find policies that maintain task-centric representations in memory, they do not try to explicitly minimize the memory. In practice, such approaches often choose the dimension of the memory with little to no knowledge of the appropriate size for the task. 

\textbf{Memory-Efficient Representations in RL.} Recent work in RL utilizes \textit{self-attention}~\citep{Vaswani17} before recurrent memory layers. For example, Baker et al. \cite{Baker19} use this method in their policy architecture to train agents to play hide-and-seek games over a long time horizon. In work by Tang et al.~\citep{Tang20}, they highlight the value of self-attention for memory-efficient representations. Specifically, they show how self-attention can be used as a bottleneck to promote the memory representation to only use task-centric features. They also demonstrate that such a bottleneck allows them to only use a small number of memory dimensions, e.g., an LSTM with only 16 memory state dimensions in a third-person perspective navigation task. While this type of approach is capable of finding task-centric representations that are low-dimensional, the memory dimension still needs to be specified \textit{a priori}. In contrast, we present a regularization scheme that explicitly seeks to minimize the memory dimension. 

A different line of work learns task-centric memory representations via information bottlenecks~\citep{Achille18, Pacelli20}. These approaches seek policies with ``low complexity"  as defined in terms of the \textit{information} contained in the memory representation. For example, in work by Pacelli and Majumdar~\cite{Pacelli20}, the objective is to minimize the information content about the state in the memory representation. Our work, instead, defines memory complexity in terms of the dimension; such a measure of complexity is more physically meaningful and tied to the robotic system's onboard memory constraints. 

\textbf{Dimensionality Reduction Techniques.} 
One approach for reducing the dimension of the memory representation is to use unsupervised techniques such as principal component analysis, manifold learning, or autoencoders~\citep{vanderMaaten09}. However, these approaches do not take into account the control task. Sufficient dimension reduction (SDR)~\citep{Adragni09} addresses this by finding a mapping of the input data such that no task-relevant information is lost. However, most existing SDR approaches are restricted to linear mappings \citep{Cook91, Li91}; the few nonlinear extensions~\citep{Kim08} rely on domain knowledge for a good kernel choice. This paper presents a method that is task-centric, handles nonlinearities, and does not require knowledge of the reduced dimension \textit{a priori}.

\section{Problem Formulation}
\label{sec:formulation}

Our goal is to find a policy that utilizes a low-dimensional, task-centric memory representation. To formalize this, we focus on robotic tasks that can be defined with cost functions of the form $\sum_{t=0}^T c_t(x_t, u_t)$ where $x_t \in \mathcal{X}, u_t \in \mathcal{U}$, and $y_t \in \mathcal{Y}$ represent the robot's state,  control action, and sensor observation at time $t$ respectively. The state space $\mathcal{X}$, action space $\mathcal{U}$, and observation space $\mathcal{Y}$ may be continuous or discrete. Additionally, the robot's dynamics and sensor model are described by unknown conditional distributions $p(x_{t+1}|x_t, u_t)$ and $s(y_t|x_t)$ respectively. 

We seek policies of the form $\pi_t(u_t|m_t)$, where $m_t$ is the memory state at time-step $t$. Here, $m_t$ is a function of the current observation and previous memory state, i.e., $m_t= q_t(y_t, m_{t-1})$. Ideally, the memory state $m_t$ should (i) contain enough information about the sequence $y_1y_2\dots y_{t-1}$ of past observations in order choose good actions, and (ii) have minimal dimension $d$, where $m_t \in \mathbb{R}^d, \forall t=0,\dots,T$. To formalize the above desiderata, we introduce the matrix zero norm\footnote{Note that, like the zero norm for vectors in Euclidean spaces, the matrix zero norm is not a proper norm because it is not homogeneous.}.

\begin{definition}[Matrix Zero Norm]
Let $a^i \in \mathbb{R}^p$ represent the transposed $i$-th row of matrix $A \in \mathbb{R}^{n \times p}$. Additionally, let
\vspace{-5pt}
\begin{align*}
    \mathbbm{1}(\|a^i\|_0 > 0) :=   \begin{cases}
                                   0, & \text{if  \ $\|a^i\|_0 = 0$} \\
                                   1, & \text{if  \ $0<\|a^i\|_0 \leq p$} 
  \end{cases}
\end{align*}
indicate if there exists a non-zero element in $a^i$. Then, the matrix zero norm is defined as the number of non-zero rows, i.e.,
\vspace{-5pt}
\begin{align*}
    \|A\|_0 := \sum_{i=1}^n \mathbbm{1}(\|a^i\|_0 > 0).
\end{align*}
\end{definition}
Thus $\|M\|_0$, where $M:=[m_0m_1\dots m_{T}]$, corresponds to the number of effective dimensions needed by the memory states across the trajectory. If $\|M\|_0 = d < D$, where $m_t \in \mathbb{R}^D$, then the memory representation is effectively reduced from dimension $D$ to $d$.

To find a memory representation that is both low-dimensional and task-centric, we minimize the memory representation dimension subject to an upper bound on the expected cost of the trajectory:
\vspace{-5pt}
\begin{equation}
\underset{\substack{q_t(y_t, m_{t-1})\\ \pi_t(u_t|m_t)}}{\text{minimize}}\ \quad \big{\|} M\big{\|}_0 \quad \text{s.t.} \quad \mathbb{E} \bigg{[} \sum_{t=0}^T c_t(x_t, u_t)\bigg{]} \leq C^{\max}, \\
\label{eqn:OPT}
\end{equation}
where $C^{\max} \in \mathbb{R}$ is the maximum allowable expected cost. Since actions are conditioned on the memory state, requiring the matrix zero norm of the memory states to be small means that the policy may need to take actions leading to lower dimensional ones. In other words, the policy actively reduces memory requirements. We call this an \textit{active memory reduction} (AMR) policy. 

\section{Learning AMR Policies}
\label{sec:AMR}
In this section, we present our approach for the AMR policy synthesis problem (\ref{eqn:OPT}). We pose the problem as a reinforcement learning problem where $q$ and $\pi$ are parameterized using neural networks. We use $w$ to refer to the combined set of weights corresponding to $q$ and $\pi$. The primary challenges with \eqref{eqn:OPT} then come from (i) the non-differentiability of the matrix zero norm, and (ii) the hard constraint on the expected cost. To tackle these, we relax the matrix zero norm with a regularizer used in group LASSO problems and soften the hard constraint. We discuss these steps, describe our overall policy gradient (PG)-style algorithm, and discuss memory-efficiency below.
\begin{figure}[t]
\floatbox[{\capbeside\thisfloatsetup{capbesideposition={right,bottom},capbesidewidth=5cm}}]{figure}[\FBwidth]
{\includegraphics[width=0.5\textwidth]{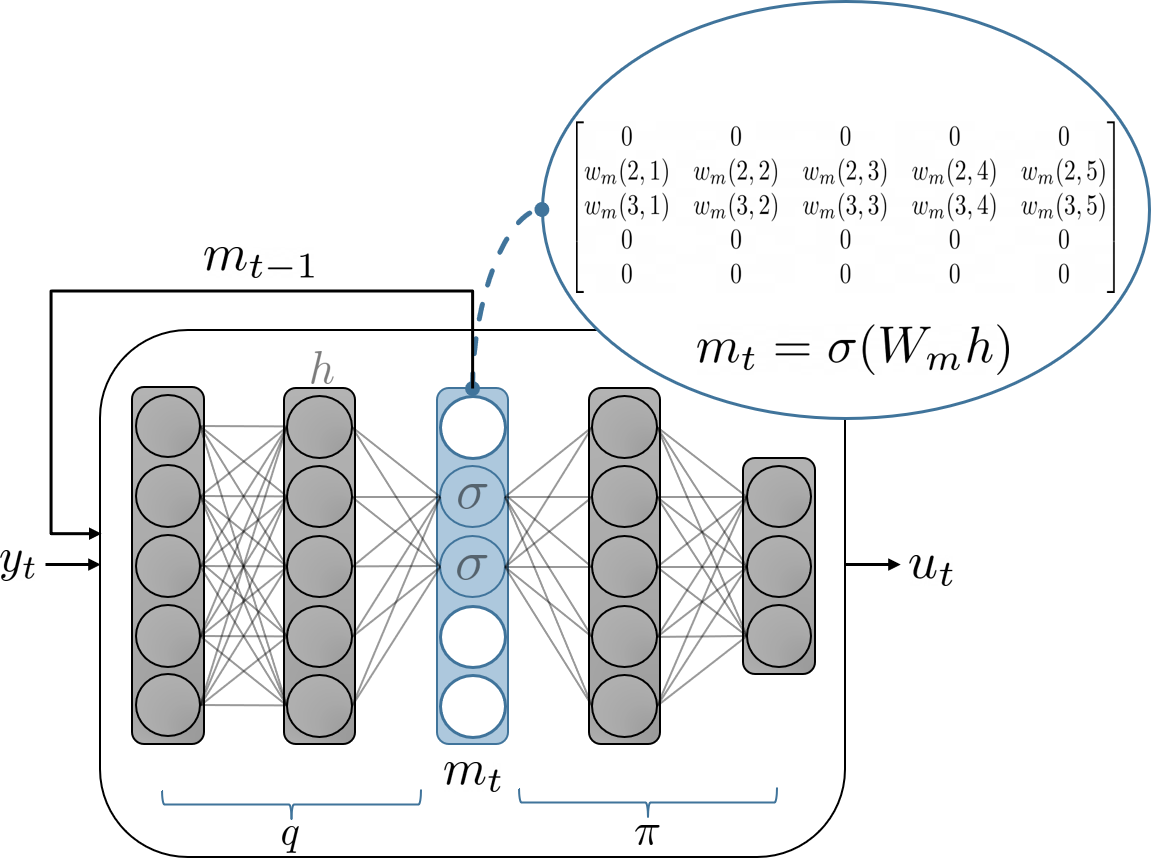}}
{\caption{\footnotesize{Effective dimensionality reduction of $m_t$ that occurs from applying the AMR regularizer defined in Section \ref{grouplasso}. The regularizer is applied to the incoming weight matrix of the memory layer (blue). The network structure shown takes $y_t$ and $m_{t-1}$ as input and outputs a distribution for $u_t$ to be sampled from.}} \label{fig:groupLASSO}}
\end{figure}
\subsection{Dimensionality Reduction Based on the $l_{2,1}$-norm}
\label{grouplasso}
A well-known and widely-used convex relaxation of the vector zero norm is the $l_1$-norm. However, since (\ref{eqn:OPT}) aims for sparsity of entire matrix rows, this relaxation cannot be used directly. Instead, we use the $l_{2,1}$-norm seen in group LASSO~\citep{Yuan06} to capture this desired behavior. The $l_{2,1}$-norm for matrix $A \in \mathbb{R}^{n \times p}$ is written as:
\begin{align}
    \|A\|_{2,1} := \sum_{i=1}^n \|a^i\|_2.
    \label{eqn:l21}
\end{align}
Notice that for $A \in \mathbb{R}^{n \times 1}$, (\ref{eqn:l21}) is the $l_1$-norm. Hence, we can expect that minimizing the $l_{2,1}$-norm will promote sparsity of entire matrix rows similar to how minimizing the $l_1$-norm promotes sparsity of elements in a vector.

The $l_{2,1}$-norm is also effective as a regularizer on groups of weights in neural networks~\citep{Scardapane17} and for promoting sparsity of hidden states in LSTMs~\citep{Wen18}. Our insight is to now apply it in an RL context to learn memory-efficient policies. First notice, for the time-invariant case, that $\Vert M \Vert_0$ is equivalent to the matrix zero norm of the incoming weights at the memory layer as shown in Figure \ref{fig:groupLASSO}. Here, ``memory layer" refers to the last layer of a standard RNN, $q^w(y_t, m_{t-1})$, that gives output $m_t$. More formally, let $d_m$ and $d_h$ be the number of neurons at the memory layer and preceding hidden layer respectively and define the incoming memory layer weight matrix to be $W_m \in \mathbb{R}^{d_m \times d_h}$. We then relax the matrix zero norm with the $l_{2,1}$-norm, i.e., $\|W_m\|_{2,1}$. Intuitively, minimizing this will promote entire rows of $W_m$ to be sparse which in turn, effectively drops out neurons (i.e., dimensions of $m_t$). 

After we relax the matrix zero norm, we soften the hard constraint on the expected cost. Our new reinforcement learning objective then becomes: 
\begin{align}
        \underset{w}{\text{minimize}}\ J(w) := \mathbb{E} \bigg{[} \sum_{t=0}^T c_t(x_t, u_t)\bigg{]} + \lambda \Big{\|} W_m\Big{\|}_{2,1}
        \label{eqn:learnAMR}
\end{align}
where $\lambda \in \mathbb{R_+}$ is a tradeoff parameter between cost and memory efficiency and can be interpreted as the inverse of the Lagrange multiplier. 

The regularizer, which we refer to as the AMR regularizer, can additionally be used in time-varying recurrent neural network structures. We achieve this by stacking the memory layer weight matrices to define $\bar W_m := [W_{m_0}, W_{m_1}, \dots, W_{m_{T}}]$ and penalizing $\|\bar W_m\|_{2,1}$. This ensures that the same number of memory dimensions are reduced at each time step.

\subsection{AMR Policy Gradient Algorithm}
\label{algorithm}

Now we describe the algorithm we use to tackle (\ref{eqn:learnAMR}). We first parameterize our policy using RNN, $q^w(y_t, m_{t-1})$, connected to a feedforward network that outputs $\pi^w(u_t|m_t)$. The output is treated as a distribution that the control action, $u_t$, is sampled from. This architecture is shown in Figure \ref{fig:groupLASSO}.

Next, we write the gradient of (\ref{eqn:learnAMR}) with respect to the network parameters, $w$, as 
\begin{align}
    \nabla_{w} J(w) = \mathbb{E}\Bigg{[} \bigg{(}\sum_{t=0}^T \nabla_{w} \log\pi^w\Big{(}u_t|q^w(y_t, m_{t-1})\Big{)} \bigg{)}\bigg{(}\sum_{t=0}^T c_t(x_t, u_t) \bigg{)}\Bigg{]}+ \lambda \nabla_w\Big{\|}W_m\Big{\|}_{2,1}.
    \label{eqn:grad}
\end{align}

In this form, we extend the canonical policy gradient algorithm, REINFORCE \citep{Williams92}, to include the AMR regularizer. 
We refer to our method as AMR-PG and outline it in  Algorithm \ref{alg:AMRPG}. For network parameter updates, we use the ADAM optimizer~\citep{Kingma14}.

\begin{algorithm}[h]
\caption{Active Memory Reduction Policy Gradient (AMR-PG)}
\begin{algorithmic}[1]
	\Repeat
		\State Rollout $N$ trajectories $\{(x_t^n, u_t^n)_{t=0}^T\}_{n=0}^{N-1}$ sampled using $\pi^w\big{(}u_t|q^w(y_t, m_{t-1})\big{)}$\ 
		
		\State $\nabla_w J(w) \approx \sum_n\big{(}\sum_t \nabla_w \log\pi^w\big{(}u_t^n|q^w(y_t^n, m_{t-1}^n)\big{)} \big{)}\big{(}\sum_t c_t(x_t^n, u_t^n) \big{)} + \lambda \nabla_w \big{\|}W_m\big{\|}_{2,1}$\ 

		\State $w \leftarrow w - \alpha \nabla_w J(w)$
	\Until Convergence of $J$
	\end{algorithmic}
	\label{alg:AMRPG}
\end{algorithm}

Once we train a policy using AMR-PG, it remains to determine the reduced memory representation dimension. In our networks used in Section \ref{sec:results}, we apply a $\tanh$ nonlinearity to the outputs preceding the memory layer. This allows us to upper bound the value of the memory state at dimension $i$ with the sum of the magnitudes of the incoming weights at dimension $i$, i.e., $m_t(i) \leq \sum_{j=1}^{d_h} |w(i,j)|$ --- we refer to the value of the upper bound as the ``memory saliency". Thus as a general rule, we cut off any dimensions whose memory saliency is at least two orders of magnitude smaller than the highest memory saliency. After determining the dimension reduction, the network can be trained for several epochs with the cut dimensions and regularizer removed. This will provide a hard dimensionality reduction if desired, i.e., explicitly force all incoming weights at a dimension to be zero. In our results discussed in Section \ref{sec:results}, we test with the raw trained network to give qualitative insight for how well the memory representation reduced its dependency on task-irrelevant features. 

\subsection{Discussion on Memory Usage and Reduction}
\label{memdiscussion}
It is important to discuss the efficacy of our approach for reducing a policy's memory requirements. In our approach, we specifically choose to focus on minimizing the RNN memory dimension as this directly determines the complexity of the memory representation for the task (motivated by Blum and Kozen's \cite{Blum78} $O(\log n)$ representation versus a map's $O(n^2)$). At initial glance, though, it may seem that the size of the network, i.e., size of RNN $q^w$ and policy $\pi^w$, can be arbitrarily large (in terms of number of layers and number of neurons/layer). 
This could potentially outweigh memory reductions achieved at the memory layer.  However, there are significant memory savings when viewing our approach holistically and applying it in a principled manner. Specifically, consider a simple RNN that has memory size: (dimension of the robot's observations) $\times$ (time horizon). Such a network avoids preemptively losing task-centric information, since the network has the space to copy each observation to memory (i.e., maintain all information the robot has received). 
As the time horizon increases (a large horizon is realistic for robotic tasks), the size of the memory layer becomes a key contributor to the overall size of the network. Thus, reducing the dimension of this layer becomes very impactful for reducing the overall memory requirements. If the overall size of the network is a concern, one can apply additional regularization as done by Scardapane et al.~\cite{Scardapane17}. However, we emphasize that this is a different objective than ours; we specifically reduce the complexity of the memory representation needed for the task. 

While applying our regularizer to the aforementioned network (with dimension: (dimension of the robot's observations) $\times$ (time horizon)) is the principled method for finding a minimal dimension, task-centric memory representation, that network may be impractically large in practice. The network designer may instead choose to make the dimension of the memory to be approximately several times (e.g., 2-5x) greater than the observation dimension; this allows for the potential to store several complete observations in memory if needed for the task. This approach is typical for standard RNN design in RL since there is little to no knowledge of what the appropriate memory dimension is.

\section{Examples}
\label{sec:results}
Here we illustrate the efficacy of our AMR-PG algorithm described in Section \ref{sec:AMR} with three examples: (i) an illustrative discrete navigation problem, (ii) a continuous navigation problem with synthetic environments, and (iii) vision-based navigation in an apartment using iGibson, a photo-realistic simulator~\citep{Xia20}. In these examples, we show that AMR-PG significantly reduces the dimension of the memory representation, improves generalization, and finds qualitatively different policies (i.e., policies that actively reduce memory) as compared to policy gradient\footnote{In these examples, the policy gradient baseline is not intended to have pre-minimized memory. Rather, the maximum RNN memory size, $D$, is chosen as it is in practice and as described in Section \ref{memdiscussion}. To verify the dimensionality reduction achieved by AMR-PG, one potential baseline is to perform a binary search for the memory size. However, this takes a prohibitively long time as it requires $O(\log D)$ full training runs (whereas AMR-PG accomplishes this in one).} with the same parameterizations. Details regarding the networks and training procedures are discussed for each example in Appendix \ref{appendix1:train}.

\begin{figure}[h]
\begin{subfigure}{0.4\textwidth}
\includegraphics[width=0.88\columnwidth, height=3.5cm]{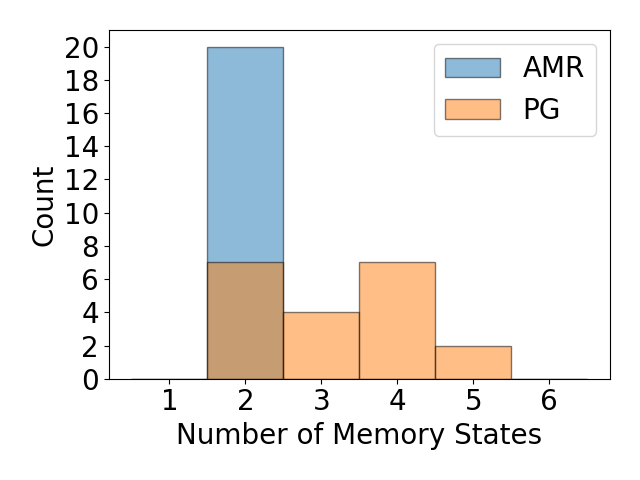}
\caption{}
\label{fig:discrete_results}
\end{subfigure}
\vspace{5mm}
\begin{subfigure}{0.3\textwidth}
\includegraphics[width=0.88\columnwidth, height=3.5cm]{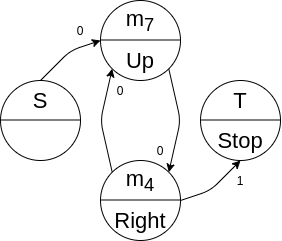}
\caption{}
\label{fig:MM}
\end{subfigure}
\begin{subfigure}{0.25\textwidth}
\includegraphics[width=0.88\columnwidth, height=3.5cm]{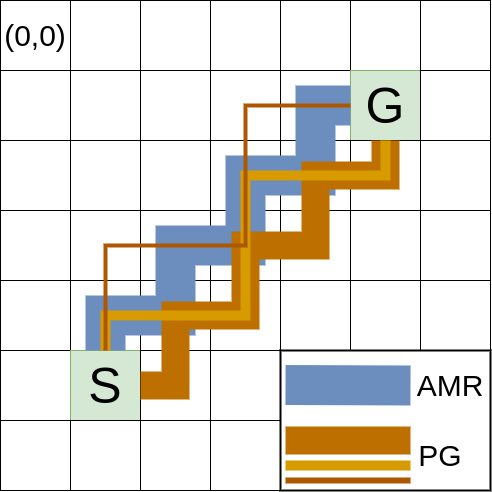}
\caption{}
\label{fig:emptymaze}
\end{subfigure}

\caption{\footnotesize{Discrete navigation results. \textbf{(a)} Number of memory states needed by AMR-PG and PG policies across 20 seeds. \textbf{(b)} A sample memory-optimal Moore machine recovered by AMR-PG where S and T are start and terminal states respectively and $m_4$ and $m_7$ are the two memory states used by the AMR-PG policy. The transitions are the robot's observations as described in Section \ref{discrete}. \textbf{(c)} The paths taken by AMR-PG (blue) and PG (various orange) policies. The weight of the line indicates the frequency of the path across the 20 seeds.}\label{fig:discrete}}

\vspace{-20pt}
\end{figure}

\subsection{Discrete Navigation}
\label{discrete}
In this first example, we specialize our method to discrete spaces in order to illustrate policies that achieve AMR. Specifically, we consider an illustrative example from~\cite{OKane17}, where a robot must navigate to a goal location in a grid as shown in Figure \ref{fig:emptymaze}. The robot is equipped with a goal indicator, e.g., $y_t=1$ means that the robot is at the goal. The robot's state, $x_t$, is described by its cell position. Additionally, the robot takes discrete actions $u_t \in \{[-1,0]^T, [0, 1]^T, [1, 0]^T, [0, -1]^T, [0, 0]^T]\}$ corresponding to \texttt{up, right, down, left,} and \texttt{stop} respectively. The state evolves with dynamics $x_{t+1}=x_t + u_t$. The cost for this scenario is $c_t(x_t, u_t) = \|x_t - g\|_1/\|x_0 - g\|_1$ for $t=0, \dots, T$ where the robot is initialized at $x_0 = [5, 1]^T$ and must navigate to goal $g = [1, 5]^T$.

The goal here is to synthesize a policy that takes the form of a deterministic, Moore-style finite state machine as described by~\cite{OKane17} and shown in Figure \ref{fig:MM}. In this context, a memory-optimal policy is defined as one that requires the fewest number of memory states. For this task, an example of a memory-optimal policy is one that simply alternates between actions \texttt{up} and \texttt{right} until the goal is observed~\citep{OKane17}. This policy only requires two memory states (not including the starting and terminal states): one for action \texttt{up} and one for action \texttt{right}; see Figure \ref{fig:MM}. In contrast, a policy that chooses to repeat $\{$\texttt{up, up, right, right}$\}$ is more complex as it requires keeping track of how many times an action has been applied; such a policy needs at least four memory states. We demonstrate that our approach recovers the optimal two-state policy identified by \cite{OKane17}. However, our method also handles continuous state, action, and observation spaces (considered in subsequent examples).

\textbf{Training and Results.} We model the memory representation with a one-hot encoding vector that indicates which memory state the robot is using (as opposed to the continuous memory states described in Section \ref{sec:AMR}). For the memory representation mapping, $q^w(y_t, m_{t-1})$, we pass the observations to the memory layer of size 10, where 10 is the maximum number of memory states this task could have (a start state, a state for each time step, and a terminal state). We use the $\beta$-Softmax activation function on the memory layer with $\beta=100$ to encourage concentration around one explicit state for $m_t$. Then we pass $m_t$ to a fully connected layer with 5 neurons activated by a Softmax function. The output is treated as a categorical distribution that we sample the actions from.

We summarize our training results for 20 seeds in Figure \ref{fig:discrete}. To count the memory states used and recover the Moore machine, we took the argmax of the memory and action layer outputs. For each seed, PG found a cost-optimal policy to the goal (see Figure \ref{fig:emptymaze}) but required between two and five memory states. In contrast, AMR-PG always found the memory-optimal policy. 
\subsection{Maze Navigation with RGB-Depth Array}
\label{mazenav}

\begin{wrapfigure}{r}{0.45\textwidth}
\vspace{-15pt}
\includegraphics[width=1\textwidth]{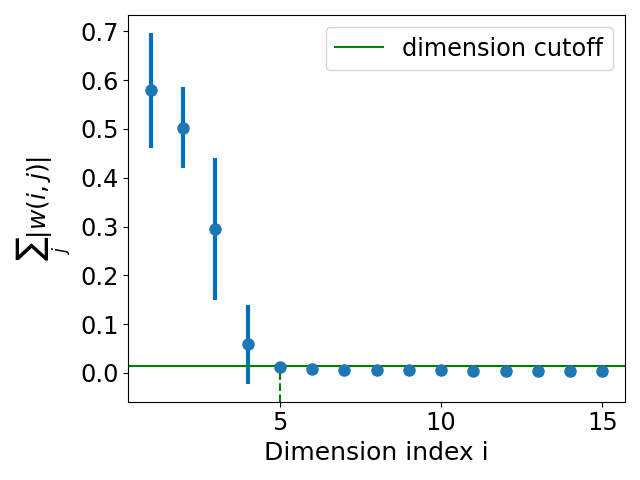}
\caption{\footnotesize{Effective memory dimension reduction for the maze example. Pictured are the top 15 memory saliency values (defined in Section \ref{algorithm}) averaged across five seeds; the error bars represent the standard deviation. }
\label{fig:memdim}
\vspace{-10pt}
}
\end{wrapfigure}

Our next example focuses on a differential-drive robot navigating through a maze (Figure~\ref{fig:cont_maze}). 
The maze is 10m $\times$ 10m with one red and one blue obstacle sampled within the shaded regions indicated in Figure \ref{fig:obstacle}. The robot is given a fixed linear velocity of 2m/s and has $T = 80$ time steps ($0.1$s each) to reach the green goal in the upper right corner of the maze. We model the cost as the Euclidean distance between the robot's position and the goal location normalized by the initial distance to the goal for all time steps. Additionally, the robot has control of its angular velocity and is equipped with a $90^\circ$ fov RGB-depth sensor that outputs colors and depths along 17 rays. The simulations are performed using Pybullet~\citep{Coumans18}.

Qualitatively, there are two policies that are sufficient for successfully navigating to the goal: (i) diagonally navigating through the obstacles to the goal, and (ii) following the maze wall to the goal. Figure \ref{fig:obstacle_test} and the video linked in Appendix \ref{appendix1:supplementary} illustrate these policies.
\begin{table}[h]
  \centering
  \begin{tabular}{|c|cc|cc|}
  	\hline
    Scenario & \multicolumn{2}{|c|}{Policy Gradient} & \multicolumn{2}{|c|}{AMR-PG}\\
    \hline
    & Cost & Dist.  & Cost & Dist. \\
    \hline
    Training & \textbf{35.09}$\pm$ \textbf{3.22} & \textbf{0.11}$\pm$\textbf{0.05} & 41.95$\pm$2.86  &  0.17$\pm$0.11 \\
    \hline
    Testing & \textbf{34.99}$\pm$ \textbf{3.25} & \textbf{0.11}$\pm$\textbf{0.04} & 42.71$\pm$3.36  & 0.19$\pm$0.13 \\
    \hline
    Testing (Swapped Colors) & 52.44$\pm$5.77 & 0.53$\pm$0.20  & \textbf{43.93}$\pm$\textbf{3.35} & \textbf{0.21$\pm$} \textbf{0.14} \\
    \hline
    Testing (New Colors) & 51.03$\pm$4.24 & 0.53$\pm$0.12  & \textbf{43.35}$\pm$\textbf{3.38} & \textbf{0.21}$\pm$\textbf{0.14} \\
    \hline
  \end{tabular}
  \vspace{-5pt}
  \caption{\footnotesize{Average costs and final normalized distances to the goal across five seeds. We used a fixed training set of 250 mazes and the same 20 testing mazes across three testing instances: (i) same color scheme used in training, (ii) swapped blue and red obstacle colors, and (iii) unseen colors on the walls and obstacles (the specific colors used are shown in Figure \ref{fig:obstacle_test}).}
  \label{tab:mazenav}
  \vspace{-10pt}}
\end{table}

\textbf{Results.}  
We compare AMR-PG with a standard PG method
that uses the same neural network parameterization (hyperparameters are provided in the Appendix). The average cost and final normalized distance to the goal (across five seeds) for training and testing scenarios are summarized in Table \ref{tab:mazenav}.  For four out of five seeds, PG found the cost-optimal solution of diagonally navigating through the obstacles. (The other seed found the wall-following policy as a result of minimal exploration outside of the far left portion of the maze). In contrast, AMR-PG consistently found the wall-following policy and \textit{significantly reduced} the required memory dimension from 300 to at most 4 as shown in Figure \ref{fig:memdim}. Thus, the policy found by AMR-PG only utilizes at most 1.33\% of the memory used by the policy found using PG. We further evaluate the benefits in terms of generalization afforded by our approach. In particular, we test the policies on environments with obstacle colors that differ from ones seen during training. The performance of the PG policies degraded significantly. In contrast, the performance of the policies found using AMR-PG remained almost entirely unaffected. This result combined with the compact memory representation suggests that AMR-PG finds policies for this problem that actively reduce memory and only maintain task-centric representations that utilize the distance values to the wall.

\subsection{Vision-Based Navigation}
\begin{figure}[h]
\begin{subfigure}{0.52\textwidth}
\includegraphics[width=1.0\linewidth]{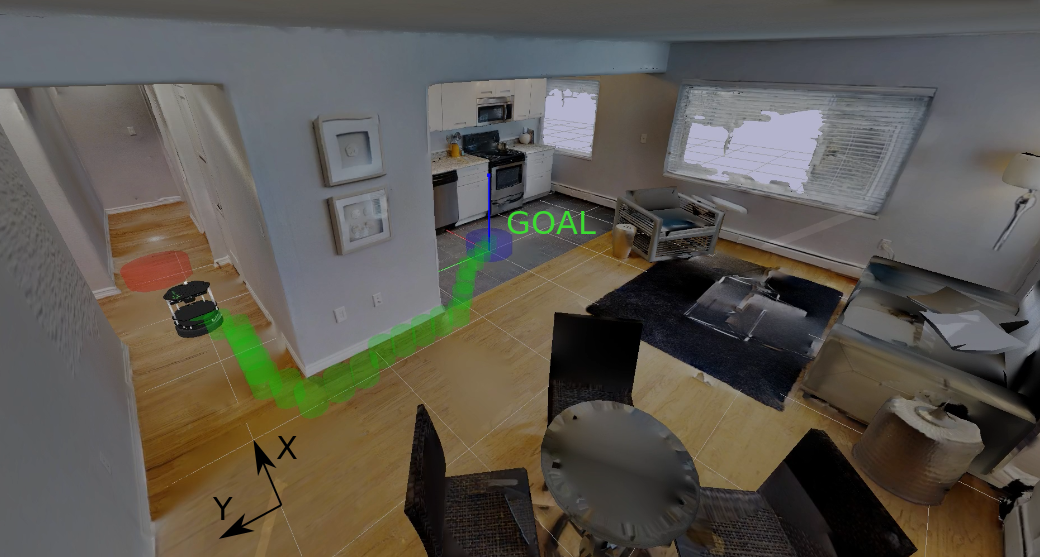}
\caption{}
\label{fig:gibsonapt}
\end{subfigure}
\begin{subfigure}{0.47\textwidth}
\includegraphics[width=1.0\linewidth]{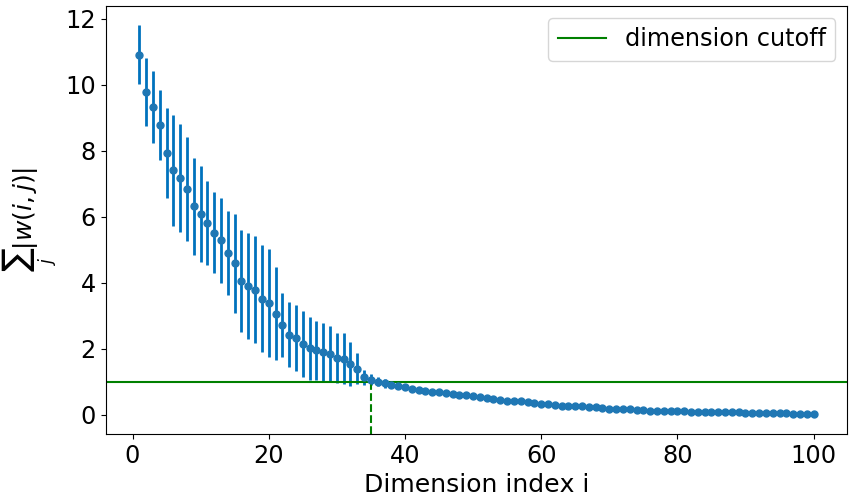}
\caption{}
\label{fig:gibsdim}
\end{subfigure}
\caption{\footnotesize{\textbf{(a)} Vision-based navigation in Placida apartment in iGibson~\citep{Xia20}. The TurtleBot is randomly initialized in the hallway (red circle) and must navigate to the kitchen (blue circle). The green path visualizes the robot's shortest path to the goal. Both AMR-PG and PG found policies that roughly follow this path. \textbf{(b)} Dimensionality reduction achieved by AMR-PG for this task (vision-based navigation). Averaged across five seeds, AMR-PG is able to reduce 100 dimensions down to 34; the error bars represent the standard deviation. }}
\label{fig:gibsonenv}
\end{figure}

The goal of our last example is to demonstrate AMR-PG's ability to scale to a more realistic scenario: vision-based navigation in a photo-realistic simulation environment. In this example, a TurtleBot is randomly initialized  in the hallway of the Placida apartment in iGibson~\citep{Xia20} and needs to navigate to the kitchen as shown in Figure \ref{fig:gibsonapt}. Specifically, the TurtleBot's initial $x$ position is sampled uniformly between the set $[x_0-2\text{m},x_0+2\text{m}]$ while the $y$ position and yaw are fixed such that the TurtleBot is centered in the hallway and facing the tables. The cost is described by a weighted sum of a sparse goal reward of 100, a term that rewards progress towards the goal (as measured by geodesic distance), a collision cost, and an angle (yaw) cost. The control actions specify linear and angular velocities. Additionally, the TurtleBot is equipped with a 90$^\circ$ fov RGB-D camera with a resolution of 128$\times$128. We preprocess these observations with a convolutional neural network before passing them to our AMR policy network. For more details, see Appendix \ref{appendix1:gibson}.

\textbf{Results.} For this example, AMR-PG found a significant memory reduction from 100 dimensions down to 34 consistently across five seeds --- a memory savings of 66\% (see Figure \ref{fig:gibsdim}). Importantly, these savings did not impact the performance of the policy. On average, AMR-PG obtained a reward of $263.7\pm34.1$ on 20 initial states sampled from the same $x$ range seen in training, while PG obtained a similar reward of $257.6\pm49.4$ on the same initial states. We also initialized the robot from 20 states drawn from an enlarged set of initial conditions: $[x_0-2.5\text{m},x_0+2.5\text{m}] \times [y_0-0.1\text{m},y_0+0.1\text{m}] \times [\theta_0-40^\circ, \theta_0+40^\circ]$.  In this case, the TurtleBot only collided once using the AMR-PG policies from the five seeds. The policies found using PG resulted in five collisions (using the same set of initial states). Thus, the policies found by AMR-PG achieve effective dimensionality reduction and show potential for improved generalization across different initial conditions. We refer the reader to the video of these results.

\section{Conclusion}
\label{sec:conclusion}
We presented a reinforcement learning approach for jointly synthesizing a low-dimensional memory representation and a policy for a given task. This joint synthesis allows one to find policies that \textit{actively} seek to reduce memory requirements. The key insight of our approach is to leverage the group LASSO regularization to encourage drop-out of neurons at the memory layer while simultaneously finding policies via a policy gradient approach. We refer to this new algorithm as AMR-PG. Additionally, we demonstrate our approach on discrete and continuous navigation problems, including vision-based navigation in a photorealistic simulator. Comparing AMR-PG and standard PG, we demonstrate that our approach can find low-dimensional representations, improve generalization, and find qualitatively different policies.

\textbf{Future Work.} There are several interesting future directions for this work. One immediate extension is to find AMR policies with actor-critic architectures (e.g., using PPO~\citep{Schulman17}), and more complex memory network architectures (e.g., LSTMs~\citep{Hochreiter97}). On the practical front, we are excited to work towards the employment of AMR policies on resource-constrained robotic platforms such as micro aerial vehicles. An important step for this is demonstrating that the AMR policies scale well to long-horizon tasks. Another potential step is to explore the benefits of pairing AMR policies with recent advances in integrated circuits that address memory accessing bottlenecks for neural networks (e.g.,~\citep{Zhang17}). Lastly, a particularly exciting direction is to explore whether our approach leads to policies that are more \textit{interpretable} (since they only maintain low-dimensional memory representations) by visualizing features that impact the memory representation (e.g., using saliency maps~\citep{Simonyan13}). 



\clearpage
\acknowledgments{This work is partially supported by the National Science Foundation [IIS-1755038], and the School of Engineering and Applied Science at Princeton University through the generosity of William Addy ’82.}


\bibliography{references}  

\newpage
\section*{\Large{Appendix}}
\label{sec:appendix}
\appendix
\section{Training Summaries}
\label{appendix1:train}

\subsection{Training Parameters}
\label{appendix1:discretetrain}
The hyperparameters used in the examples are detailed in Table \ref{tab:appendixtrain}.
\begin{table}[h]
  \centering
  \begin{tabular}{| m{8em} | m{1.5cm}| m{1.5cm}| m{1.5cm}| m{1.5cm}| m{1.5cm}|}
  	\hline
    Example & Learning Rate & AMR Rate ($\lambda$) & Max Memory Dimension & Max Epochs & \# Rollouts\\
    \hline
    \hline
    Discrete Nav. & \num{1e-1} & \num{1e-1}  & 10 &  300 & 100 \\
    \hline
    Maze Nav. & \num{1e-4} & \num{5e-1} & 300 & 3000 &  250 \\
    \hline
    Vision-Based Nav. & \num{1e-4} & \num{5e-1} & 100 & 6000  & 3 \\
    \hline
  \end{tabular}
  \caption{Hyperparameters used in training.}
  \label{tab:appendixtrain}
  \vspace{-5pt}
\end{table}

\subsection{Maze Navigation with RGB-Depth Array Example}
\label{appendix1:maze}
Here, we describe our network implementation for the maze navigation example. For the input to the neural network, we feed in a flattened RGB-depth sensor observation (along 17 rays) so that $y_t \in \mathbb{R}^{68}$. We additionally augment $y_t$ with the previous action so that $\tilde y_t = [y_t, u_{t-1}]^T$. The recurrent network, $q^w(\tilde y_{t}, m_{t-1})$, has two hidden layers with 369 neurons each with exponential linear unit and $\tanh$ nonlinearities respectively. The memory output layer is 300 and also has a $\tanh$ nonlinearity. This is then fully connected to the output layer with 2 neurons. For turning rate control, we treat the 2-dimensional network output as a Gaussian distribution. 

\subsection{Vision-Based Navigation Example}
\label{appendix1:gibson}
Before $q^w(y_{t}, m_{t-1})$, we preprocess the image and depth sensor observations with a convolutional neural network containing three layers using filter sizes 32, 64, and 64, kernel sizes 8, 4, and 3, and strides 4, 2, and 2 respectively. The output is then fully connected to a layer of size 256 activated by a $\tanh$ nonlinearity. This is then treated as $y_t$ and passed to the recurrent network, $q^w(y_t, m_{t-1})$, with a memory of size 100. Similar to the maze navigation example, we pass the memory state, $m_t$, to a fully connected network whose output is treated as a multivariate Gaussian distribution for applying the linear and angular velocities. For implementing AMR-PG, we modify the REINFORCE agent in the TF-Agent Tensorflow library~\citep{TFAgents}.

In this example, we use a weighted sum of various reward types including a sparse goal reward, a geodesic potential, a collision cost, and a yaw cost. The weights are given in Table \ref{tab:appendixweights}.  We model the geodesic potential as the previous geodesic distance minus the current geodesic distance. The collision cost is a penalty applied to any time step with a collision. The yaw cost is also applied at each time step and describes the angular difference between the TurtleBot's yaw angle and the angle needed to get to the goal. We use this cost to promote the TurtleBot to move forward to the goal.
\begin{table}[h]
  \centering
  \begin{tabular}{| m{8em} | m{1.5cm}| m{1.5cm}| m{1.5cm}| m{1.5cm}| m{1.5cm}|}
  	\hline
     & Sparse Goal & Geodesic Potential & Collision & Yaw \\
    \hline
    \hline
    Weight & 100 & 30  & -0.5 &  -0.2  \\
    \hline
  \end{tabular}
  \caption{Reward weights.}
  \label{tab:appendixweights}
  \vspace{-5pt}
\end{table}

\section{Supplementary Material}
\label{appendix1:supplementary}
A video of our results is available at: \url{https://youtu.be/x5yYhLoG6jY}

Our code is available at: \url{https://github.com/irom-lab/AMR-Policies}

\end{document}